\pdfoutput=1

\documentclass[11pt]{article}

\usepackage[preprint]{acl}

\usepackage{times}
\usepackage{latexsym}

\usepackage[T1]{fontenc}
\usepackage[utf8]{inputenc}
\usepackage{microtype}
\usepackage{inconsolata}
\usepackage{graphicx}
\usepackage{amsmath}

\usepackage{amssymb}

\usepackage{algorithmic,algorithm}
\renewcommand{\algorithmiccomment}[1]{\bgroup\hfill$\triangleright$~#1\egroup}
\usepackage[linguistics]{forest} 
\usepackage{synttree}
\usepackage{subcaption}
\usepackage{tikz-dependency}
\usepackage[cjk]{kotex}

\usepackage{booktabs}
\usepackage{langsci-gb4e}

\newcommand*{\ycc}[1]{{\color{black}#1}}
\newcommand*{\yc}[1]{{\color{black}#1}}

\title{Better Heads Do Not Guarantee Better Binarized Constituency Parsing}

\author{
Zeyao Qi\raisebox{-0.1\height}{\includegraphics[height=0.7em]{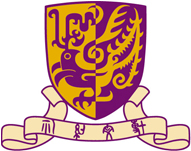}}~~~ 
Yige Chen\raisebox{-0.1\height}{\includegraphics[height=0.7em]{CUHK.jpg}}~~~ 
Eitan Klinger\raisebox{-0.1\height}{\includegraphics[height=0.7em]{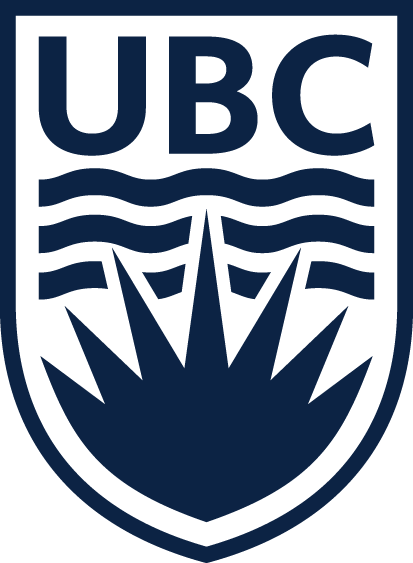}}~~~ 
Vivaan Wadhwa\raisebox{-0.1\height}{\includegraphics[height=0.7em]{UBC.png}}~~~
Jungyeul Park\raisebox{-0.1\height}{\includegraphics[height=0.7em]{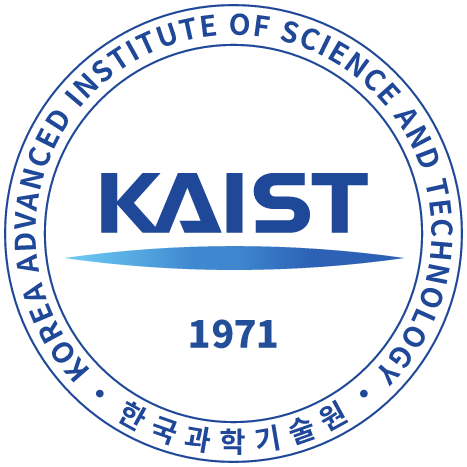}}\thanks{~~Corresponding author.}\\
\raisebox{-0.1\height}{\includegraphics[height=0.9em]{CUHK.jpg}} The Chinese University of Hong Kong, Ma Liu Shui, Hong Kong\\ 
\raisebox{-0.1\height}{\includegraphics[height=0.9em]{UBC.png}} The University of British Columbia, Vancouver, Canada\\
\raisebox{-0.1\height}{\includegraphics[height=0.9em]{KAIST}} Korea Advanced Institute of Science \& Technology, Daejeon, South Korea\\
\url{https://ling.cuhk.edu.hk}~~~ \url{https://www.cs.ubc.ca}~~~ \url{https://ct.kaist.ac.kr}
\\}

\begin{document}
\maketitle

\begin{abstract}
We revisit punctuation-aware tree binarization for constituency parsing and ask whether dependency-induced headedness improves binary parser supervision. Although learned heads substantially outperform rule-based heads in intrinsic head prediction, they do not yield consistent parsing gains after debinarization. In particular, punctuation-conditioned evaluation shows that learned headedness underperforms rule-based binarization in macro-average punctuation-sensitive $F_1$, despite a small overall gain on CTB. Similar instability appears under cross-treebank transfer. These results suggest that \ycc{linguistically grounded} headedness is not necessarily parser-optimal when used as a binarization control signal. The paper presents a negative result: better head prediction does not imply better punctuation-sensitive constituency parsing.
\end{abstract}

\section{Introduction}

Tree binarization converts non-binary constituency trees into binary trees so that parsing algorithms can operate over uniform local branching decisions.
It is central to chart-based parsing, including CKY-style algorithms \citep{cocke:1969,kasami:1966,younger:1967}, and it has also shaped transition-based constituency parsing, where parser actions are often defined over binary or incrementally binarized structures \citep{sagae-lavie-2005-classifier,zhu-etal-2013-fast}.
Because binarization is reversible, it is often treated as a technical preprocessing step whose effects disappear after debinarization.

This view is incomplete when binarized trees serve as parser training targets.
A binary tree does not only encode the recoverable non-binary tree; it also specifies the order of local composition, the intermediate categories exposed to the learner, and the structural domains over which parsing decisions are trained.
Different binary trees may debinarize to the same original tree while presenting different supervision signals.
Binarization is therefore not merely an encoding convention, but an interface between treebank annotation and parsing architecture.

This interface is especially consequential for head-driven binarization.
The selected head child anchors the binary spine of a constituent and determines how non-head material is composed around it.
Since most constituency treebanks do not annotate head children, head-driven binarization has traditionally relied on hand-written head percolation rules \citep{collins-1999-head}, which encode treebank-specific projection conventions.

A natural alternative is to induce heads from aligned dependency annotation.
For each constituency span, the head child can be defined as the child that dominates the dependency head of that span.
This yields a linguistically grounded and independently evaluable head source.
Recent work has also reconsidered head selection in neural constituency parsing by treating heads as dynamic or latent parser-internal decisions \citep{hou-li-2025-dynamic}.
Our question is different: if heads are used not as latent model variables, but as control signals for constructing binary parser targets, does more accurate headedness yield better parsing supervision?

We test this question under punctuation-aware binarization.
Punctuation is often ignored or removed under PARSEVAL and standard \texttt{evalb} conventions \citep{black-etal-1991-procedure}, although it marks clause boundaries, appositive domains, coordination breaks, parentheticals, quotations, and sentence closure.
Removing punctuation before binarization changes the local configurations from which binary trees are derived.
We therefore construct binary targets that retain punctuation as boundary structure before binary branching is imposed, and compare rule-based heads with dependency-induced learned heads under otherwise identical parser training and debinarization conditions.

The motivating hypothesis is straightforward.
If punctuation supplies boundary information and dependency-induced heads better approximate syntactic projection, then punctuation-aware binarization with learned heads should provide better binary supervision.
Our results do not support this hypothesis.
Although learned heads substantially outperform rule-based heads in intrinsic head prediction, they do not yield consistent downstream parsing gains after debinarization.
The learned-head condition is slightly better on CTB overall, but not on PTB, and this small gain disappears under punctuation-conditioned evaluation.
\ycc{There}, learned headedness underperforms rule-based binarization in macro-average punctuation-sensitive $F_1$ on both treebanks.
Similar instability appears under cross-treebank transfer.

%This negative result is the main contribution of the paper.
\yc{The paper's main contribution is a negative result}: head accuracy is not binarization utility.
The child that best matches an independently annotated dependency analysis is not necessarily the child that yields the most learnable binary target for a constituency parser.
\ycc{Linguistically grounded headedness} may make the binarization interface more explicit, but it is not automatically parser-optimal.

\section{Tree binarization with punctuation}
\label{sec:punct-binarization}

We define the binarization procedure used to construct parser targets from constituency trees whose terminals include lexical tokens and punctuation.
Let $T$ be such a tree, and let $H$ specify the head child of each nonterminal node.
The function $f_{\mathrm{punct}}(T;H)$ maps $T$ to a binary tree $T^{\mathrm{bin}}$ in two steps: punctuation is first retained in the local child sequence as boundary structure, and the resulting sequence is then binarized by head-driven composition.
Thus, punctuation is present when binary structure is constructed, rather than removed before binarization and reattached afterward.

The transformation is reversible; \yc{formally,} $f^{-1}_{\mathrm{punct}}(f_{\mathrm{punct}}(T;H)) = T$.
The inverse removes only intermediate binarization nodes and restores the original ordered children, including punctuation.
Punctuation-aware binarization therefore defines a reversible parser target, not a new annotation layer.

\paragraph{Punctuation as boundary structure}

Punctuation is treated as boundary structure, not as a phrasal head.
It constrains the local domain in which non-punctuation material is grouped.
Right-boundary punctuation, such as commas and periods, is grouped with the material to its left:
$
  (X \; \alpha \; \beta \; p)
  \Rightarrow
  (X \; (\texttt{@X} \; \alpha \; \beta) \; p),
$
where $p$ is right-boundary punctuation.
Left-boundary punctuation, such as opening quotation marks and opening parentheses, is treated dually:
$
  (X \; p \; \alpha \; \beta)
  \Rightarrow
  (X \; p \; (\texttt{@X} \; \alpha \; \beta)).
$
The $\texttt{@X}$ node marks an intermediate constituent introduced by the transformation and records the boundary position of punctuation without treating punctuation as the projecting head.

\paragraph{Head-driven binary conversion}

After boundary positions have been fixed, the full child sequence is binarized.
For a nonterminal node $v$ with label $X$ and ordered children $\mathit{ch}(v)=\langle v_1,\ldots,v_k\rangle$, let $h(v)=v_j$ be its designated head child.
The conversion introduces $\texttt{@X}$ intermediate nodes until the local branching factor is binary, while preserving $v_j$ as the projection anchor of $X$.
Because $h(v)$ is selected over the full child sequence, punctuation remains part of the local configuration used to construct the binary spine.

\paragraph{Reversibility}

Debinarization removes only the $\texttt{@}$-prefixed intermediate nodes and restores the original ordered children: $f^{-1}_{\mathrm{punct}}(T^{\mathrm{bin}})=T$.
Labels, spans, terminal order, and punctuation tokens are preserved.
Recoverability alone does not determine parser supervision, since multiple binary trees may debinarize to the same original tree.
Punctuation-aware binarization therefore requires punctuation to shape binary construction, not merely to survive debinarization.

\paragraph{Head sources}

We compare two head sources only as binarization controls.
The rule-based condition uses Collins-style percolation rules.
The learned condition uses a previously trained dependency-induced head selector, which predicts the child containing the dependency head of each constituent span.
Head prediction itself is not a contribution of this paper; it is used only to test whether intrinsically stronger headedness yields better binary parser targets.

\section{Experiments}
\label{sec:experiments}

We evaluate whether intrinsically stronger heads yield better binary parser targets under punctuation-aware binarization.

\paragraph{Intrinsic head accuracy}
\label{sec:head-finding-experiments}

Table~\ref{tab:head-accuracy} shows that the learned \yc{head} selector substantially outperforms rule-based heads against dependency-induced gold heads.
These heads are used only as binarization controls in the parsing experiments.\footnote{$\mathcal{R}$ denotes rule-based head finding, and $\mathcal{L}$ denotes learned head finding.}

\begin{table}[!ht]
\centering
\footnotesize
\begin{tabular}{r cc}
\toprule
 & $\mathcal{R}$ & $\mathcal{L}$ \\
\midrule
PTB \citep{marcus-etal-1993-building} & 67.89 & 98.54 \\
CTB \citep{xue-etal-2005-ctb} & 92.42 & 100.00 \\
\bottomrule
\end{tabular}
\caption{Intrinsic head prediction accuracy against dependency-induced gold heads.}
\label{tab:head-accuracy}
\end{table}

\paragraph{Parsing with punctuation-aware binarization}
\label{sec:parsing-binarization}

For each head source, we binarize the training trees, train the same parser, and debinarize predicted trees before evaluation against the original gold trees.
Thus, the parser and evaluation target are fixed; only the binary supervision changes.
Table~\ref{tab:parsing-accuracy} reports labeled bracket \(F_1\).\footnote{$\mathcal{R}_{\text{\o}}$ denotes rule-based binarization without punctuation awareness, $\mathcal{R}_{\text{p}}$ rule-based punctuation-aware binarization, and $\mathcal{L}_{\text{p}}$ learned-head punctuation-aware binarization.
\texttt{evalb} excludes punctuation from evaluation \citep{black-etal-1991-procedure}, whereas \textsc{jp-evalb} retains punctuation-bearing constituents \citep{jo-etal-2024-novel}.}

\begin{table}[!ht]
\centering
\footnotesize
\begin{tabular}{c ccc}
\toprule
&  & \textsc{evalb} & \textsc{jp-evalb} \\
\midrule
PTB 
 & $\mathcal{R}_{\text{\o}}$ & 95.50 & 94.92 \\
 & $\mathcal{R}_{\text{p}}$ & 95.97 & 95.43 \\
 & $\mathcal{L}_{\text{p}}$ & 95.88 & 95.35 \\
\midrule
CTB 
 & $\mathcal{R}_{\text{\o}}$ & 92.17 & 90.29 \\
 & $\mathcal{R}_{\text{p}}$ & 94.26 & 92.40 \\
 & $\mathcal{L}_{\text{p}}$ & 94.46 & 92.56\\
\bottomrule
\end{tabular}
\caption{Parsing accuracy after deterministic debinarization.}
\label{tab:parsing-accuracy}
\end{table}

Despite the large intrinsic advantage in Table~\ref{tab:head-accuracy}, learned heads do not yield consistent gains after debinarization.
They are slightly below rule-based punctuation-aware binarization on PTB and slightly above it on CTB.
The overall scores therefore show that punctuation-aware targets remain trainable, but they do not show that better head prediction reliably gives better binary parser supervision.

\paragraph{Punctuation-sensitive evaluation}
\label{sec:punctuation-evaluation}

Table~\ref{tab:punctuation-summary} reports macro-average punctuation-conditioned \(F_1\) with \textsc{jp-evalb}; full punctuation-by-punctuation scores and gold counts appear in Appendix~\ref{app:detailed-punctuation-evaluation}.
The learned-head condition \yc{underperforms} both rule-based alternatives on both treebanks.

\begin{table}[!ht]
\centering
\footnotesize
\begin{tabular}{lccc}
\toprule
& $\mathcal{R}_{\text{\o}}$ 
& $\mathcal{R}_{\text{p}}$ 
& $\mathcal{L}_{\text{p}}$ \\
\midrule
PTB & 92.63 & 92.43 & 92.27 \\
CTB & 94.22 & 93.99 & 93.50 \\
\bottomrule
\end{tabular}
\caption{Macro-average punctuation-conditioned \(F_1\) scores with \textsc{jp-evalb}.}
\label{tab:punctuation-summary}
\end{table}

\paragraph{Cross-treebank binarization}
\label{sec:cross-treebank-binarization}

We further test whether transferred heads can construct punctuation-aware parser targets when target-specific head supervision is unavailable.
A head selector trained on a source treebank is applied to a target treebank after deterministic label normalization; the predicted heads are then used only to anchor the binary spine for punctuation-aware binarization.
Parser training and evaluation follow the same debinarization-based protocol as above.
This setting is intended to test whether head selection can be reused across resources, rather than whether transferred headedness is intrinsically optimal.

\begin{table}[!ht]
\centering
\footnotesize
\begin{tabular}{r ccc}
\toprule
 & $\mathcal{R}_{\text{p}}$ & $\mathcal{T}_{\text{p}}$ & $\mathcal{L}_{\text{p}}$ \\
\midrule
Sinica \citep{huang-etal-2000-sinica} & N/A & 79.03 & N/A \\
FTB \citep{abeille-clement-toussenel:2003} & 88.65 & 86.82 & 88.92 \\
\bottomrule
\end{tabular}
\caption{Cross-treebank parsing accuracy under punctuation-aware binarization \yc{with \textsc{jp-evalb}}. Transferred heads are predicted by a selector trained on CTB; Appendix~\ref{app:transfer-details} details the transfer procedure. 
% Scores are labeled bracket \(F_1\) after deterministic debinarization and are computed with \textsc{jp-evalb}.
}
\label{tab:transfer-parsing}
\end{table}

Table~\ref{tab:transfer-parsing} shows that transferred heads \((\mathcal{T}_{\text{p}})\) can provide usable binary targets, but they do not reliably match target-specific or resource-specific head sources.
The degradation on FTB indicates that projection conventions remain treebank-sensitive even when the binarization procedure itself is fixed.
This reinforces the main result: \ycc{head selection is not a neutral preprocessing choice, but a control signal whose utility depends on the target annotation scheme.}

\section{Analysis}
\label{sec:analysis}

The experiments separate punctuation preservation, linguistic headedness, and parser-optimal binarization.
Punctuation is syntactically meaningful, learned heads are more faithful to dependency projection, and the transformation is reversible.
Yet these properties do not yield better punctuation-sensitive parsing.
The negative result is therefore specific: better head prediction does not imply better binary supervision.

This does not weaken the linguistic status of punctuation.
Earlier work showed that grammars retaining punctuation can outperform those that discard it \citep{jones-1994-exploring}, and that punctuation contributes to syntactic analysis \citep{briscoe-carroll-1995-developing,jones-1996-towards-testing}.
Nunberg's account of punctuation as a structured subsystem remains central here \citep{nunberg-1990-linguistics}.
Our result instead narrows the computational claim: punctuation should remain visible, but its presence does not determine which head source yields the most learnable binary target.

\paragraph{Head accuracy and binarization utility}

A dependency-induced head identifies the child containing the dependency head of a constituent span.
A binarization head determines how that constituent's children are recursively grouped into a binary training target.
These notions are related, but they optimize different objects.
Intrinsic head evaluation rewards agreement with dependency projection; parser training rewards binary structures that are regular, learnable, and recover accurate constituent spans after debinarization.

Reversibility hides this distinction.
Many binary trees can collapse to the same original constituent.
Head choice may therefore change the intermediate composition order without changing the final bracket set.
Since labeled bracket \(F_1\) is computed after debinarization, it cannot reward a head source merely for being linguistically more faithful.
The parser learns from intermediate structure, but evaluation largely removes it.

\paragraph{Punctuation and boundary-sensitive composition}

Punctuation-bearing constituents expose the mismatch because punctuation is boundary-sensitive rather than head-like.
Commas, periods, quotation marks, parentheses, colons, and semicolons mark closure, interruption, apposition, quotation, coordination-like separation, parenthetical insertion, and sentence boundary.
Their contribution is configurational: punctuation shapes the domain in which other material is composed.

Dependency heads do not necessarily encode this boundary function.
In dependency annotation, punctuation is often attached to nearby lexical material or handled by conventions for dependency well-formedness.
In constituency annotation, the same punctuation mark may delimit a larger phrasal or clausal domain.
A learned head can therefore be correct for dependency projection while producing a binary spine that is less favorable for punctuation-sensitive constituency parsing.

This explains the punctuation-conditioned degradation.
Rule-based heads impose stable category-based composition patterns.
Learned heads introduce dependency-aligned variation into the binary spine, and around punctuation this variation is not consistently useful.
\yc{The result does not imply that punctuation is irrelevant. Instead, it shows that punctuation-sensitive parsing depends on the interaction between boundary structure and learnable composition order, not on head accuracy alone.}

\paragraph{Binarization as a parser interface}
Punctuation-aware binarization is a parser interface, not a transparent encoding.
Punctuation defines boundary-sensitive local domains; headedness selects projection anchors inside those domains.
These roles cannot be collapsed into a single notion of linguistic correctness.
A head source may be intrinsically better yet less useful as a binarization control signal.
The choice of head source is therefore a modeling decision, mediated by learnability, regularity, and compatibility with the parser objective.

\section{Conclusion}

This paper \yc{presents} a negative result for punctuation-aware binarization: learned heads that are substantially more accurate under dependency-induced head evaluation do not yield reliably better parser targets after debinarization. They give a small overall gain on CTB, but not on PTB, and underperform rule-based alternatives in macro-average punctuation-sensitive evaluation on both treebanks. The result does not diminish the syntactic relevance of punctuation; rather, it shows that punctuation-sensitive parsing cannot be reduced to better headedness. Punctuation defines boundary-sensitive composition domains, while headedness selects projection anchors within them. These dimensions interact, but they are not interchangeable: a dependency-motivated head may still induce a binary spine that is less regular, less learnable, or less aligned with punctuation-bearing constituency structure.

The broader implication is that binarization is not a transparent encoding of treebank structure. It is an interface through which annotation decisions become parser supervision. Reversibility guarantees recoverability, but not utility; linguistic explicitness guarantees interpretability, but not parser optimality. A parser-optimal binarization may therefore require punctuation-specific composition rules, parser-aware regularization, or direct optimization of binary targets, rather than reliance on intrinsically accurate head prediction alone.

\section*{Limitations}

The learned heads used in our experiments are dependency-induced, but the dependencies are not fully independent of constituency annotation.
For English and Chinese, the dependency representations are obtained from treebank conversion or closely related annotation pipelines.
The learned heads should therefore not be interpreted as heads from \yc{an entirely} separate syntactic theory.
They provide an independently evaluable projection signal, but one still shaped in part by constituency-to-dependency conversion assumptions.

The evaluation is limited to a fixed parsing architecture.
The negative result concerns the use of learned heads as binarization control signals under this parser and evaluation setting.
Other parsers, especially models with different inductive biases over binary structure, may interact differently with the same binarized targets.

Punctuation-by-punctuation evaluation is sensitive to the distribution of punctuation marks.
Frequent marks such as commas and periods provide relatively stable estimates, whereas rare marks yield less reliable scores.
The analysis should therefore be read as evidence of uneven punctuation-sensitive behavior, not as a definitive claim about each individual punctuation symbol.

Finally, the punctuation-aware procedure uses a finite punctuation inventory and deterministic boundary conventions.
This makes the transformation explicit and reversible, but it may not capture all uses of punctuation, especially in quotation, apposition, parenthetical insertion, and discourse-level segmentation.
A more expressive approach may require punctuation-specific composition rules or parser-aware optimization of binary targets.

%%\bibliography{references}

\appendix

\section{Background and motivation}
\label{sec:background}

\paragraph{Binarization as a representational interface}
\label{sec:binarization-interface}

Constituency treebanks often contain flat or variably branching structures, whereas many parsing algorithms operate over binary branching trees.
Binarization resolves this mismatch by replacing each non-binary constituent with a sequence of binary combinations.
Although this transformation is reversible, its role changes when binarized trees are used as parser targets.

In that setting, binarization specifies an order of local composition.
It determines which daughters combine first, which intermediate projections are introduced, and which local structural contrasts are exposed to the learner.
Two binary trees may debinarize to the same original constituent while encoding different composition histories.
Binarization is therefore a representational interface between treebank annotation and parser supervision, not only a lossless encoding convention.

\paragraph{Punctuation exclusion in parser preprocessing}
\label{sec:punctuation-exclusion}

Parser pipelines often inherit punctuation-exclusion conventions from PARSEVAL-style evaluation and standard \texttt{evalb}, where punctuation is typically omitted from labeled bracket scoring \citep{black-etal-1991-procedure}.
For evaluation, this reduces sensitivity to annotation-specific punctuation treatment.
For binarization, however, the same convention changes the structural input to the transformation.
If punctuation is removed before binarization, boundary-bearing tokens are absent precisely when local binary structure is constructed.

Punctuation marks are not merely orthographic residue.
They delimit clause boundaries, appositive domains, parentheticals, quotations, coordination breaks, and sentence closure.
Subsequent reinsertion may recover the terminal string, but it cannot recover the same composition history.
This distinction becomes visible under punctuation-sensitive evaluation such as \textsc{jp-evalb}, which retains punctuation-bearing constituents during scoring \citep{jo-etal-2024-novel}.

\begin{figure}[!ht]
\centering
\subfloat[Original tree with punctuation \label{original-tree}]{
\footnotesize{
\begin{forest}
for tree={
  align=center,
  rounded corners,
  if n children=0{
    tier=word
  }{}
}
[S [NP [\textit{The}] [\textit{little}] [\textit{boy}]]
[VP [\textit{likes}] [NP [\textit{red}] [\textit{tomatoes}]]]
[$\cdot$]]
\end{forest}
}}
\quad
\subfloat[Standard binarization \label{wrong-binary}]{
\footnotesize{
\begin{forest}
for tree={
  align=center,
  rounded corners,
  if n children=0{
    tier=word
  }{}
}
[S [NP [\textit{The}] [@NP [\textit{little}] [\textit{boy}]]]
[@S [VP [\textit{likes}] [NP [\textit{red}] [\textit{tomatoes}]]] [$\cdot$]]]    
\end{forest}
}}
\quad
\subfloat[Punctuation-aware binarization \label{correct-binary}]{
\footnotesize{
\begin{forest}
for tree={
  align=center,
  rounded corners,
  if n children=0{
    tier=word
  }{}
}
[S [@S [NP [\textit{The}] [@NP [\textit{little}] [\textit{boy}]]]
[VP [\textit{likes}] [NP [\textit{red}] [\textit{tomatoes}]]]] [$\cdot$]]
\end{forest}
}}
\caption{Standard binarization may construct binary structure after punctuation has been excluded, whereas punctuation-aware binarization preserves the boundary relation before binary branching is imposed.}
\label{treebank-binary-tree-example}
\end{figure}
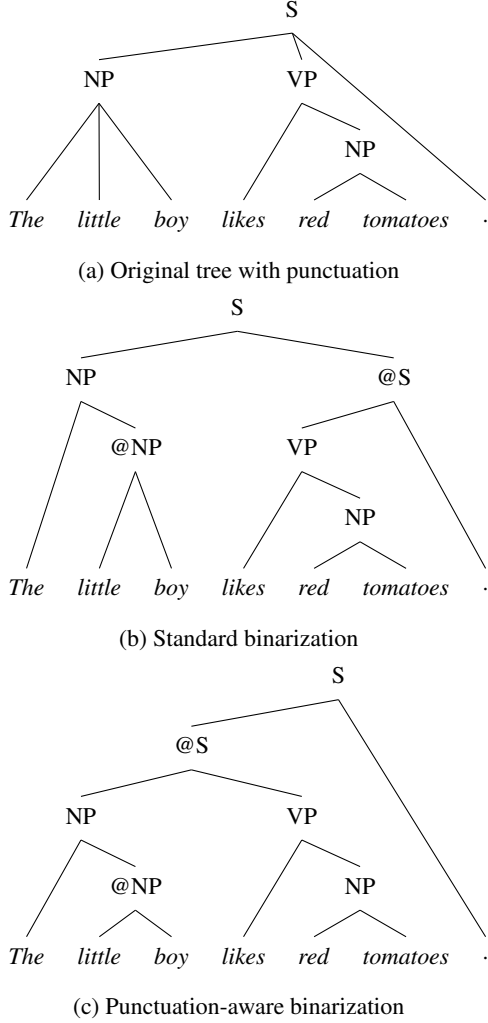

Figure~\ref{treebank-binary-tree-example} illustrates the contrast.
A standard pipeline that excludes punctuation during binarization may later attach it near the root, yielding the structure in Figure~\ref{wrong-binary}.
The punctuation-aware alternative in Figure~\ref{correct-binary} keeps punctuation inside the local constituent configuration before binary branching is imposed.

\paragraph{Head selection and projection conventions}
\label{sec:head-rules}

Head-driven binarization adds a further representational choice.
When a constituent has more than two children, the selected head child anchors the binary spine and determines how non-head material is composed around it.
Because most constituency treebanks do not annotate head children, this choice is usually supplied by hand-written percolation tables, such as Collins-style rules for the Penn Treebank \citep{collins-1999-head}.
Such tables encode projection conventions that may be useful, but they need not match dependency-induced heads or transfer across treebanks.

Dependency-induced heads provide an explicit alternative.
Given aligned constituency and dependency trees, the head child of a constituency node can be defined as the child that dominates the dependency head of the corresponding span.
This makes head selection independently evaluable.
The experiments in the main text ask whether this intrinsic advantage also improves binary parser supervision.

\paragraph{Why reversibility is not enough}
\label{sec:reversibility}

Reversibility is necessary but not sufficient.
A binarized tree must debinarize to the original constituency tree, \ycc{but many binary trees satisfy this condition}.
They collapse to the same non-binary structure while presenting different intermediate categories, attachment orders, and local decision contexts during training.

This distinction matters because parsers are trained on binary trees, not on their debinarized equivalence classes.
A reversible binarization can preserve the final tree while changing the supervision signal.
Punctuation exclusion and head-source mismatch \yc{illustrate this issue}: the former removes boundary-bearing tokens before local composition is determined, while the latter supplies projection anchors that may not be optimal for parser learning.

\section{Head sources for binarization}
\label{app:learning-heads}

This appendix describes the head sources used by \(f_{\mathrm{punct}}(T;H)\).
Headedness is not treated as an independent parsing task in this paper.
For each nonterminal node, a head child is selected only to anchor the binary spine used as a parser training target.

\subsection{Dependency-induced head supervision}
\label{app:gold-heads}

Let \(T=(V,E)\) be a rooted constituency tree with terminals \(w_1,\ldots,w_n\), and let \(V_{\mathrm{NT}}\) be the set of nonterminal nodes.
For each \(v\in V_{\mathrm{NT}}\), let
\[
  \mathit{ch}(v)=\langle v_1,\ldots,v_k\rangle
\]
be its ordered children.
A headed constituency tree assigns one head child to each nonterminal node:
\[
  h(v)\in \mathit{ch}(v).
\]

Head supervision is induced from aligned dependency annotation.
Let \(D\) be a dependency tree over terminals \(\{1,\ldots,n\}\), with artificial root \(0\), and let
\[
  g:\{1,\ldots,n\}\rightarrow\{0,1,\ldots,n\}
\]
be the dependency governor function.
For a nonterminal node \(v\), let \(\mathit{yield}(v)\) be the set of terminals dominated by \(v\).
The dependency span-head candidates are:
\[
H(v)=
\{\, i\in \mathit{yield}(v) : g(i)\notin \mathit{yield}(v) \,\}.
\]
If \(H(v)\) contains exactly one token, that token is taken as the dependency head of the span.
The gold head child is the unique immediate child of \(v\) that dominates this token.
Nodes for which \(H(v)\) is empty or non-singleton are excluded from supervised head learning and intrinsic evaluation.

This definition identifies the child that makes a constituency span compatible with the aligned dependency analysis.
It supplies the head information required for head-driven binarization.

\subsection{Head prediction model}
\label{app:head-model}

Each training instance is a nonterminal node.
The input is the parent label and the ordered sequence of child labels:
\[
  \ell(v),\ \ell(v_1),\ldots,\ell(v_k),
\]
and the target is the head-child index \(j\) such that \(h(v)=v_j\).
The model is unlexicalized and receives no word forms, lexical embeddings, or dependency features.
Phrase labels and part-of-speech tags are normalized by removing redundant functional or morphological suffixes.

The label sequence is encoded with a Transformer encoder \citep{devlin-etal-2019-bert}.
A multilayer perceptron scores each child position, and the highest-scoring child is selected as the head.
The model is trained with cross-entropy loss over valid child positions.
The rule-based baseline uses Collins-style head percolation rules \citep{collins-1999-head}, as implemented in Stanford CoreNLP for English and Chinese.

\subsection{Head transfer}
\label{app:transfer-details}

For cross-treebank transfer, a selector trained on a source resource \(\mathcal{R}_s\) is applied to a target resource \(\mathcal{R}_t\).
Because treebanks differ in phrase labels, part-of-speech tags, and annotation conventions, target labels are normalized into the source label space with deterministic mappings:
\[
\phi:\Sigma^{\mathrm{ph}}_t\rightarrow\Sigma^{\mathrm{ph}}_s,
\qquad
\psi:\Sigma^{\mathrm{pos}}_t\rightarrow\Sigma^{\mathrm{pos}}_s.
\]
The selector predicts a head-child index for the normalized target configuration, and the predicted index is interpreted in the original target tree.

Table~\ref{tab:transfer-heads} reports intrinsic transfer accuracy.
The results show partial transfer across resources, with CTB \(\rightarrow\) Sinica at 69.41\%, PTB \(\rightarrow\) FTB at 54.95\%, and CTB \(\rightarrow\) FTB at 64.70\%.
Transfer therefore depends not only on language identity, but also on compatibility between source and target projection conventions.

\begin{table}[!ht]
\centering
\footnotesize
\begin{tabular}{r c}
\toprule
Transfer setting & Head accuracy \\
\midrule
CTB \(\rightarrow\) Sinica & 69.41 \\
PTB \(\rightarrow\) FTB & 54.95 \\
CTB \(\rightarrow\) FTB & 64.70 \\
\bottomrule
\end{tabular}
\caption{Intrinsic cross-treebank and cross-lingual transfer accuracy of learned headedness.}
\label{tab:transfer-heads}
\end{table}

The main error source in PTB \(\rightarrow\) FTB is PP headedness.
Under the dependency-induced criterion, PTB PPs are overwhelmingly head-final, while FTB and CTB PPs are almost categorically head-initial.
This convention mismatch explains why PTB \(\rightarrow\) FTB transfers poorly, despite the apparent structural similarity of the resources.

\section{Additional experimental details}
\label{app:experimental-details}

\subsection{Data and alignment}
\label{app:data-alignment}

The main experiments use the Penn Treebank~3 \citep{marcus-etal-1993-building} and the Penn Chinese Treebank~5.1 \citep{xue-etal-2005-ctb}.
For English, we use Sections~02--21 for training, Section~22 for development, and Section~23 for testing.
For Chinese, we follow the standard CTB~5.1 constituency parsing split.\footnote{\url{https://github.com/nikitakit/self-attentive-parser/tree/master/data/ctb_5.1}}

For head learning, constituency and dependency trees are aligned at the terminal level.
After normalization, we retain only sentences whose constituency and dependency representations have identical terminal sequences.
This ensures that head supervision is induced only from tokenization-consistent structures.
The same splits are used for parser training and evaluation.

\subsection{Parser training and evaluation}
\label{app:parser-training}

We use the Berkeley Neural Parser \citep{kitaev-klein-2018-constituency,kitaev-cao-klein:2019:ACL}.
Rather than using its default binarization, we provide punctuation-aware binary trees as parser targets.

For each treebank, the original constituency trees are transformed under each head source:
\[
\begin{aligned}
\mathcal{T}^{\mathrm{bin}}_{\mathrm{rule}}
&=
f_{\mathrm{punct}}\!\left(\mathcal{T};H_{\mathrm{rule}}\right),\\
\mathcal{T}^{\mathrm{bin}}_{\mathrm{learn}}
&=
f_{\mathrm{punct}}\!\left(\mathcal{T};H_{\mathrm{learn}}\right).
\end{aligned}
\]
The parser architecture, training configuration, intermediate-node convention, punctuation treatment, and debinarization procedure are identical across conditions.
Only the head source used to construct the binary targets differs.

At test time, predicted binary trees are deterministically debinarized before evaluation against the original unbinarized gold trees.
Thus, differences in parsing accuracy reflect differences in binary supervision, not differences in the final evaluation target.
We report labeled bracket \(F_1\) with both standard \texttt{evalb} and \textsc{jp-evalb} \citep{jo-etal-2024-novel}, the latter retaining punctuation during constituent evaluation.

\subsection{Punctuation-sensitive evaluation}
\label{app:detailed-punctuation-evaluation}

For punctuation-sensitive analysis, we restrict evaluation to constituents containing a given punctuation mark and compute \(F_1\) with \textsc{jp-evalb}.
Table~\ref{tab:punctuation-results} reports punctuation-conditioned scores for each mark.
The row \(\mathcal{G}\) gives the corresponding number of gold punctuation-bearing constituents.
The macro-average columns are summarized in Table~\ref{tab:punctuation-summary}.

\begin{table*}
\centering
\subfloat[PTB \label{ptb-punct}]{
\footnotesize
\resizebox{\textwidth}{!}{
\begin{tabular}{l ccc ccc ccc ccc ccc ccc cc}
\toprule
 & , & \$ & . & `` & '' & ' & \texttt{-RRB-} & \texttt{-LRB-} & ; & : & \texttt{-LCB-} & \texttt{-RCB-} & \# & ... & ` & ? & - & = & ! & \textsc{avg}\\
\midrule
$\mathcal{R}_{\text{\o}}$ & 91.38 & 92.77 & 97.35 & 92.15 & 91.99 & 95.42 & 89.89 & 90.32 & 94.29 & 88.19 & 92.68 & 92.68 & 95.56 & 87.64 & 87.72 & 74.07 & 93.75 & 83.33 & 100.00  & 92.63\\
$\mathcal{R}_{\text{p}}$ & 91.65 & 92.65 & 97.47 & 92.39 & 91.70 & 96.12 & 91.08 & 91.52 & 95.38 & 90.63 & 90.16 & 90.16 & 97.78 & 88.89 & 93.10 & 76.36 & 93.75 & 66.67 & 66.67 & 92.43\\
$\mathcal{L}_{\text{p}}$ & 91.37 & 92.57 & 97.51 & 91.81 & 91.13 & 95.42 & 88.16 & 88.45 & 93.81 & 91.12 & 90.32 & 90.32 & 97.78 & 91.11 & 88.14 & 69.09 & 93.75 & 83.33 & 0.00 & 92.27\\
\midrule
$\mathcal{G}$ & 10067 & 2822 & 2439 & 1680 & 1283 & 346 & 309 & 307 & 162 & 125 & 65 & 65 & 45 & 43 & 29 & 29 & 15 & 6 & 1 & 19,838\\
\bottomrule
\end{tabular}
}
}

\subfloat[CTB \label{ctb-punct}]{
\centering
\footnotesize
\resizebox{.75\textwidth}{!}
{
\begin{tabular}{l ccc ccc ccc ccc ccc}
\toprule
 & ， & 、 & 。 & “ & ” & （ & ） & 《 & 》 & ； & — & ： & —— & ！ & \textsc{avg}\\
\midrule
$\mathcal{R}_{\text{\o}}$ & 93.88 & 92.45 & 99.61 & 94.58 & 94.44 & 99.19 & 98.39 & 97.92 & 97.92 & 97.67 & 97.56 & 95.65 & 66.67 & 100.00 & 94.22\\
$\mathcal{R}_{\text{p}}$ & 93.89 & 92.39 & 99.61 & 94.63 & 93.92 & 99.19 & 98.39 & 92.93 & 92.93 & 97.67 & 85.71 & 95.65 & 70.59 & 100.00 & 93.99 \\
$\mathcal{L}_{\text{p}}$ & 93.03 & 91.49 & 99.61 & 95.57 & 94.97 & 97.56 & 97.56 & 98.95 & 98.95 & 93.62 & 95.24 & 95.65 & 50.00 & 100.00 & 93.50 \\
\midrule
$\mathcal{G}$ & 1359 & 1009 & 256 & 102 & 90 & 62 & 62 & 47 & 47 & 22 & 20 & 11 & 10 & 1 & 3098\\
\bottomrule
\end{tabular}
}
}
\caption{Punctuation-conditioned $F_1$ scores with \textsc{jp-evalb}.}
\label{tab:punctuation-results}
\end{table*}

\subsection{Structural comparison of binary targets}
\label{app:binary-structural-comparison}

To measure how much the head source changes parser supervision before debinarization, we compare binary targets directly.
For each original tree, we compare \(T^{\mathrm{bin}}_{\mathrm{rule}}\) and \(T^{\mathrm{bin}}_{\mathrm{learn}}\).

We report bracketing $F_1$ between the two binary trees and exact binary spine identity.
Bracketing $F_1$ measures overlap in intermediate spans.
Exact spine identity is stricter, requiring the full binary structure introduced for an original constituent to match.
A high bracketing score with lower spine identity indicates local rotations within the same original constituent.

\subsection{Cross-treebank head transfer}
\label{app:cross-transfer-details}

Cross-treebank transfer is used only to construct binary parser targets.
The source and target label inventories are normalized with deterministic mappings, after which the source-trained head selector is applied to the target constituency trees.
The resulting heads determine the projection anchor used by punctuation-aware binarization.
This setting applies to constituency resources that lack aligned dependencies or resource-specific head rules.

\subsection{Alignment with English CCGbank}
\label{app:ccgbank-alignment}

We compare punctuation-aware PTB binary trees with aligned CCGbank derivations \citep{hockenmaier-steedman-2007-ccgbank}.
\ycc{The comparison is conducted on Section~23 of the Penn Treebank.}
CCGbank provides derivations for 2,407 of the 2,416 original sentences.
After exact terminal alignment, 1,858 sentences remain; most exclusions involve tokenization mismatches around punctuation and quotation marks.

To focus on structural correspondence rather than category inventory, we simplify both representations before evaluation.
All nonterminal labels are replaced with \texttt{nt}, POS--word terminals are preserved, and type-raising unary rules are removed from CCGbank.
Structural overlap is then evaluated with \textsc{jp-evalb}, which retains punctuation during constituent evaluation.

The punctuation-aware PTB binary trees obtain \(F_{1}=76.07\) against the simplified CCGbank derivations.
The remaining mismatches mostly reflect different grammatical treatments of punctuation.
In appositive expressions such as \textit{the CEO, John Smith}, our binarization attaches the comma to the preceding noun phrase to preserve local constituency.
CCGbank often associates the comma with the appositive phrase and promotes it to a higher derivational position, giving it a role closer to coordination.
These divergences reflect different assumptions about punctuation structure rather than annotation noise.

\section{Additional analysis}
\label{app:additional-analysis}

\subsection{Algorithmic interpretation}
\label{app:algorithmic-interpretation}

Punctuation-aware binarization separates two operations that are easily conflated.
First, punctuation-bearing boundary configurations are fixed inside the original local child sequence.
Second, the resulting configuration is converted into a binary tree by head-driven composition.
A procedure that only groups punctuation with neighboring material is therefore not yet a binarization procedure.
Conversely, a head-driven binarization that operates after punctuation has been removed is not punctuation-aware in the relevant sense.

This distinction also explains why reversibility is too weak as a criterion.
The identity
\[
  f^{-1}_{\mathrm{punct}}(f_{\mathrm{punct}}(T;H)) = T
\]
states recoverability, but not usefulness as parser supervision.
Many reversible transformations preserve the final tree while changing the intermediate structures on which the parser is trained.
The relevant question is therefore whether punctuation participates in binary structure construction, not merely whether it can be restored after debinarization.

\subsection{Punctuation-specific instability}
\label{app:punctuation-specific-instability}

The punctuation-conditioned results are uneven because the evaluation slices are not linguistically homogeneous.
Sentence-final punctuation often marks a regular closure relation.
Commas are frequent but functionally diverse, spanning coordination, apposition, clause linkage, parentheticals, and discourse-level segmentation.
Quotation marks and brackets occur in paired constructions, where the left and right symbols have different configurational roles.
Rare punctuation marks yield less stable scores, since a small number of local errors can substantially change \(F_1\).

This heterogeneity weakens the expectation of uniform gains from a single head source.
Dependency-induced heads may help when dependency projection and constituency boundary structure coincide, but hurt when punctuation functions as a delimiter independent of the lexical head.
The punctuation-by-punctuation analysis is computed only with \textsc{jp-evalb}, whose role is to keep punctuation-bearing constituents visible to evaluation.
The observed degradation therefore indicates that punctuation-sensitive constituency structure is not reducible to headedness.

\end{document}